\documentclass{article}
\usepackage{iclr2027_conference,times}
\iclrfinalcopy
\usepackage[T1]{fontenc}
\usepackage{amsmath,amssymb,amsthm,booktabs,tabularx}
\usepackage{graphicx,float}
\usepackage[hidelinks]{hyperref}
\usepackage{microtype}
\newtheorem{theorem}{Theorem}
\newtheorem{corollary}{Corollary}
\newtheorem{proposition}{Proposition}
\newcommand{\E}{\mathbb E}
\newcommand{\Prob}{\mathbb P}
\title{Statistical G\"odel Machines:\\From Formal Proofs to Risk-Controlled Self-Improvement}

\author{
\textbf{Xuening Wu}$^{1}$,
\textbf{Shenqin Yin}$^{1}$,
\textbf{Yanlan Kang}$^{1}$,\\
\textbf{Xinhang Zhang}$^{1}$,
\textbf{Qianya Xu}$^{3}$,
\textbf{Zeping Chen}$^{2}$,
\textbf{Wenqiang Zhang}$^{1}$\\[0.5em]
\normalfont $^{1}$Fudan University\\
\normalfont $^{2}$Tongji University\\
\normalfont $^{3}$University of California, San Diego
}
\makeatletter
\renewcommand{\@maketitle}{%
  \begin{center}
    {\LARGE\scshape \@title\par}
    \vspace{1.5em}
    {\normalsize
      \begin{tabular}{c}
        \@author
      \end{tabular}\par}
    \vspace{1em}
  \end{center}
}
\makeatother

\date{}
\begin{document}
\maketitle
\fancyhead{}
\renewcommand{\headrulewidth}{0pt}
\pagestyle{fancy}
\thispagestyle{fancy}

\begin{abstract}
G\"odel machines formulate self-improvement through program modifications justified by formal proofs of increased utility. Obtaining these proofs is difficult in learning systems evaluated through noisy empirical observations. We introduce Statistical G\"odel Machines (SGM), a framework that replaces the per-update formal proof requirement with a statistical improvement certificate for a specified evaluation objective. SGM separates adaptive proposal generation from confirmation and permits a modification only when its evidence satisfies a trajectory-level risk allocation. Under conditionally valid confirmation tests and a predictable, summable error budget, the probability of ever accepting a non-improving update is bounded by a user-specified level. For a fixed utility and faithful state transitions, this also gives a high-probability non-decreasing incumbent-utility sequence, without guaranteeing that improvements will be found. The framework uses established statistical tools and makes explicit the assumptions and weaker guarantees of this alternative to formal proof search. Controlled numerical and synthetic experiments quantify erroneous commitments, retained improvements, and verification cost. A post-hoc decomposition distinguishes reductions in zero-effect commitments from reductions in strictly harmful commitments. A two-attempt ImageNet-100 run demonstrates a certified checkpoint commitment followed by a rejected candidate trained from the adopted state. This short finite-pool demonstration establishes executable adoption, not long-horizon error calibration; historical learning trajectories remain exploratory. SGM provides a practical protocol for evidence-based self-improvement, while exposing the trade-off between erroneous commitments, missed improvements, and verification cost.
\end{abstract}

\section{Introduction}
A self-improving system must decide not only which modification to propose, but also when the available evidence justifies adopting it. The G\"odel machine gives a principled answer: execute a self-modification when a formal proof establishes its usefulness relative to continuing the current process, including its proof search~\citep{schmidhuber2003godel}. This formulation accommodates uncertainty through probabilistic models. Its practical difficulty is obtaining the required model and proof for changes to complex learning systems, not an incompatibility between formal reasoning and stochastic environments.

SGM studies a statistical relaxation of this proof obligation. A candidate is evaluated against the incumbent on a specified objective, and acceptance requires a statistical certificate. The certificate tolerates a controlled probability of error and is conditional on a sampling model. It does not establish the original machine's full future-utility comparison or global optimality. This weaker claim is deliberately chosen to make evidence-based adoption executable in empirical settings.

Consider an automated learning loop that proposes a lower learning rate, trains a branch from the current checkpoint, and observes higher development accuracy. The adoption decision concerns the resulting frozen candidate: fresh confirmation must justify installing it before subsequent proposals are trained from that adopted state. Replacing the training procedure for future random executions is a different decision, requiring evidence about the procedure's expected utility. The improvement obligation must specify which of these objects is being adopted.

Accepted updates can change subsequent proposals, training states, and resource use. Although a checkpoint can often be restored, resources already spent and opportunities encountered on a different search path may not be recovered. These consequences motivate controlling erroneous commitments across the trajectory. They do not imply that every update is physically irreversible, or that classical statistical methods require decisions to be reversible.

Our objective is to bound the probability of at least one non-improving commitment, while allowing useful modifications to pass. Rejecting every proposal trivially avoids false commitments, so risk must be evaluated together with acceptance power and verification cost. SGM is independent of the proposer: it does not improve the search algorithm by itself.

\paragraph{Contributions.}
We (i) specify a statistical replacement for the formal improvement obligation and delimit its relationship to G\"odel-machine self-improvement; (ii) give an executable proposal--confirmation--commit protocol with a conditional trajectory guarantee and a fixed-objective monotonicity corollary; and (iii) quantify false commitments, retained utility, and evidence cost in controlled numerical and synthetic trajectories, distinguish their error composition through post-hoc analysis, and document a learning loop with a certified commitment, subsequent proposals from the adopted state, and a held-out audit. The probabilistic argument composes established inference tools. Confirm-Triggered Harmonic Spending (CTHS) is an optional allocation instance, rather than a prerequisite for the framework.

\begin{figure}[tbp]
\centering
\includegraphics[width=.86\linewidth]{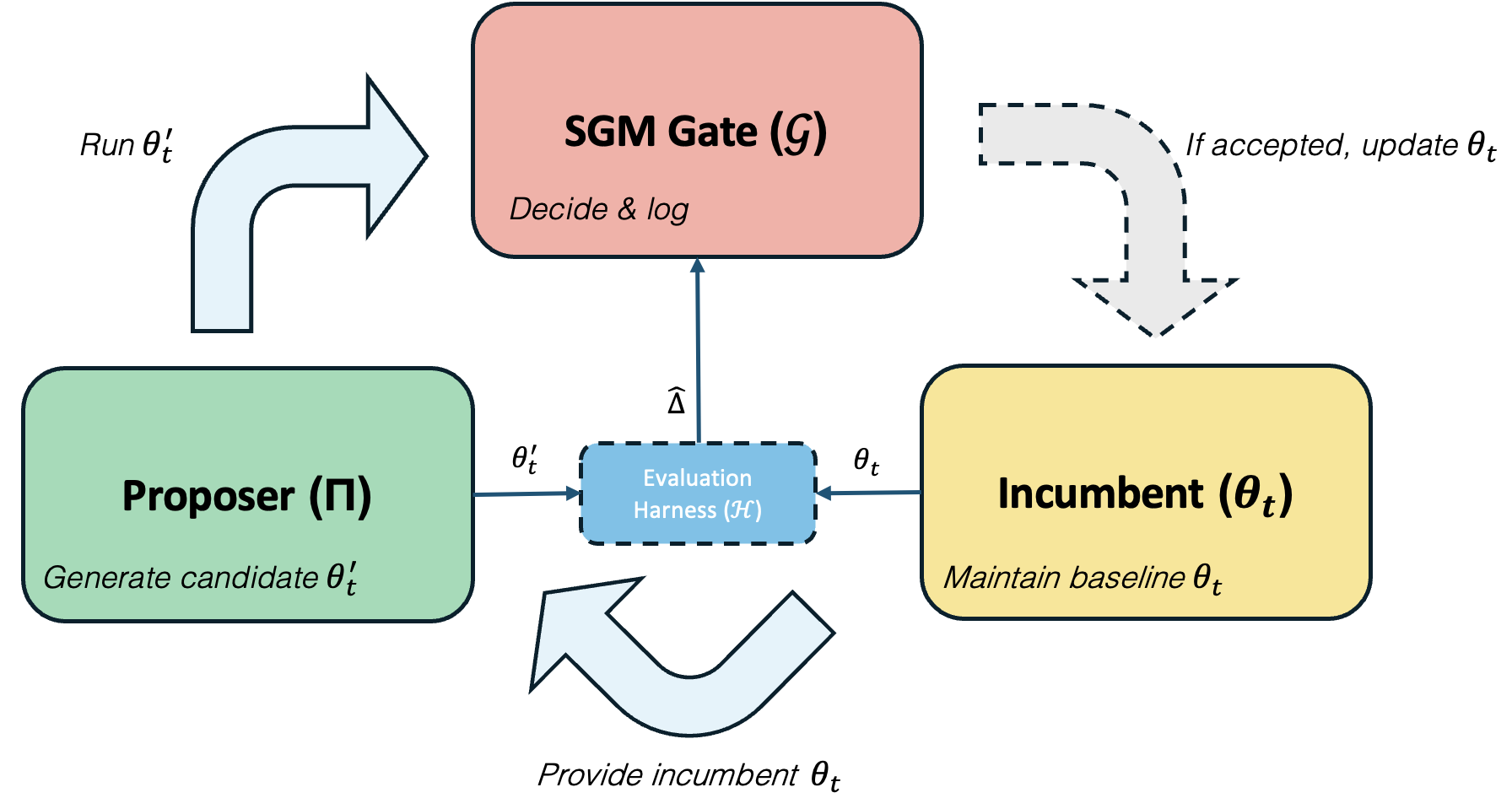}
\caption{SGM proposal--evaluation--commit architecture. The proposer supplies a candidate and the evaluation harness compares it with the incumbent. The gate authorizes adoption using a statistical certificate and the allocated error budget. The diagram uses $\theta_t$ and $\theta_t^\prime$ for the incumbent and candidate; the formal protocol indexes opened confirmation attempts by $k$.}
\label{fig:architecture}
\end{figure}

Figure~\ref{fig:architecture} summarizes the roles of the proposer, evaluation harness, and acceptance gate.

\section{Related work and scope}

G\"odel machines motivate proof-justified self-modification~\citep{schmidhuber2003godel}; Darwin G\"odel Machines use empirical evaluation to guide code-modifying search~\citep{zhang2025darwin}. The October 2025 SGM preprint~\citep{wu2025sgm} proposed statistical improvement certificates and a global error budget for recursive edits. This empirical obligation concerns a declared evaluation objective, rather than the G\"odel machine's full future-utility comparison.

\paragraph{Earlier statistical approval of model modifications.}
\citet{feng2021approval} formulate automatic algorithmic change protocols for evolving medical prediction models, using prospective data and comparisons to approved models. They study noninferiority-induced deterioration (bio-creep), beneficial approvals, and online bad-approval counts and ratios. Their infinite-window count guarantee bounds the probability of any bad approval for their error definition; it is not merely a per-update guarantee. They also identify declining approval power from alpha-wealth depletion. This is a primary antecedent, alongside online FWER control. SGM instead specifies strict improvement in a declared scalar utility against the actual incumbent and an explicit certificate-to-state execution protocol, with checkpoint and training-procedure targets distinguished. These scope choices do not establish a new statistical primitive or superiority over their methods. Neither statistical approval of persistent updates nor alpha-death is claimed as new here.

\paragraph{Subsequent certification frameworks.}
SEA~\citep{sengupta2026sea}, released in July 2026, combines anytime-valid gates and a fixed error budget with a broader agent architecture. Its SGM-CS controller explicitly builds on SGM and reuses its confirm-triggered spending. SEA distinguishes established statistical components from their endogenous, distribution-shifting composition, whose safety it leaves unproved. Safe Harness~\citep{cai2026safeharness}, released in September 2026, analyzes modification generation, finite-data certification and selection, safe adoption, and opportunities after an update. It derives finite-run guarantees and separates generation from certification bottlenecks. Both postdate the SGM preprint; their overlapping concerns are important subsequent developments, rather than antecedents of its statistical adoption formulation.

The present treatment makes the declared utility, candidate-selection history, conditional certificate validity, and actual adopted state explicit. It gives a trajectory guarantee under predictable, summable error allocation and studies risk, retained improvement, and evidence cost in controlled trajectories. This focus complements SEA's broader architecture and Safe Harness's feasibility analysis; it does not establish superiority over either framework or guarantee future improvement opportunities. Learning case studies with unresolved certificates remain illustrative evidence.

\paragraph{Statistical foundations.}
High-confidence policy improvement provides probabilistic guarantees for policy updates and repeated improvement~\citep{thomas2015high}. Online family-wise error control handles potentially unbounded hypothesis sequences~\citep{tian2019online}; confidence sequences and betting methods support optional stopping~\citep{howard2021time,waudby2024estimating}. SGM organizes these established tools into an adoption protocol. Persistent state changes do not invalidate these foundations, and the trajectory bound is a conditional composition of them.

\paragraph{Adaptive search and resource allocation.}
Population Based Training adapts parameters and hyperparameters during training~\citep{jaderberg2017population}. Hyperband allocates resources through early stopping~\citep{li2018hyperband}, and BOHB combines model-guided search with bandit resource allocation~\citep{falkner2018bohb}. These methods can supply candidates to SGM; adoption additionally requires confirmation that remains conditionally valid after adaptive selection.

\paragraph{Protocol comparison.}
Table~\ref{tab:protocol-comparison} contrasts targets and guarantee scopes. HCPI already includes repeated policy improvement. Comparisons require matching the target, evidence access, and total error budget; we claim no empirical superiority over HCPI, SEA, or Safe Harness.

\section{A statistical improvement obligation}
\subsection{State, utility, and confirmation history}
Let $s_k$ be the incumbent before confirmation attempt $k$, and let $c_k$ be a candidate selected using all previous observations and optional development screening. Let $\mathcal G_k$ contain this pre-confirmation history, the selected states, the evaluation rule, and the allocated level $\alpha_k$. The index counts opened confirmation attempts, including failed attempts, rather than accepted changes or outer proposal rounds.

For the primary formulation fix an evaluation distribution $P$ and a bounded score $r(s,z)\in[0,1]$. Define
\[
 U(s)=\E_{Z\sim P}[r(s,Z)],\qquad
 \mu_k=U(c_k)-U(s_k),\qquad H_k=\{\mu_k\leq0\}.
\]
The true-null event $H_k$ is $\mathcal G_k$-measurable: the selected states determine their utility difference even though it is unknown to the algorithm. A false commitment is $H_k\cap A_k$, where $A_k$ is the event that the candidate is adopted. This event includes zero improvement; strict harm corresponds to $\mu_k<0$.

For example, fresh paired observations are $X_{k,i}=r(c_k,Z_{k,i})-r(s_k,Z_{k,i})\in[-1,1]$. Conditional IID sampling from $P$, after fixing the candidate, is a sufficient model. More general sampling is possible only with a justified conditional-validity argument. The target may instead concern training-seed performance on a fixed dataset; this must not be confused with a fixed model's performance on new examples.

\paragraph{The object of certification.}
The observation unit must match the object being adopted (Table~\ref{tab:certificate_object}). For a frozen checkpoint, randomness can come from sampled evaluation examples. For a frozen training procedure, a paired observation instead compares scores from fresh random executions under fixed data and execution rules. A procedure certificate concerns expected utility over those executions, not every resulting checkpoint. Certifying a particular trained checkpoint therefore does not certify its training rule across future random executions. The historical multi-seed experiments illustrate the latter evaluation setting, but their unresolved certificates are not evidence of a validated procedure-level commitment.

\begin{table}[htbp]
\centering\small
\caption{Two certification targets. The chosen object, distribution of random inputs, and scoring rule must be fixed before confirmation.}
\label{tab:certificate_object}
\begin{tabularx}{\linewidth}{@{}p{.18\linewidth}XX@{}}\toprule
Adopted object & One paired observation & Certified utility \\\midrule
Frozen checkpoint & Candidate and incumbent scores on a newly sampled evaluation example & Expected score under the declared evaluation distribution \\
Frozen training procedure & Candidate and incumbent scores from a fresh paired random execution under fixed data and scoring rules & Expected score over the declared execution randomness \\\bottomrule
\end{tabularx}
\end{table}

\subsection{Certificates and the global guarantee}
A certificate supplies a conditionally valid acceptance event:
\begin{equation}\label{eq:valid}
 \mathbf1_{H_k}\Prob(A_k\mid\mathcal G_k)
 \leq \mathbf1_{H_k}\alpha_k\quad\text{almost surely}.
\end{equation}
Here $\alpha_k$ is chosen before confirmation observations, and $\sum_k\alpha_k\leq\delta$ almost surely. Adopting a candidate also requires that the incumbent and candidate actually match the objects to which the certificate refers.

\begin{theorem}[Trajectory-level false-commit control]\label{thm:risk}
If Equation~\eqref{eq:valid} holds for every opened confirmation and the predictable allocation is pathwise summable with total at most $\delta$, then, for an adaptive sequence containing any mixture of null and improving candidates,
\[
 \Prob\!\left(\bigcup_{k\geq1}(H_k\cap A_k)\right)\leq\delta.
\]
\end{theorem}
\begin{proof}
Measurability and conditional validity give
$\Prob(H_k\cap A_k)=\E[\mathbf1_{H_k}\Prob(A_k\mid\mathcal G_k)]\leq\E[\mathbf1_{H_k}\alpha_k]$.
A countable union bound and the pathwise budget constraint yield the result. Unopened attempts have empty acceptance events. Independence between different attempts is not required; conditional validity within each selected attempt is required.
\end{proof}

\begin{corollary}[Fixed-objective self-improvement]\label{cor:mono}
Suppose the only state changes are $s_{k+1}=c_k$ after acceptance and $s_{k+1}=s_k$ otherwise, and the same $U$ is used throughout. With probability at least $1-\delta$, every accepted change strictly increases $U$ and the incumbent-utility sequence is non-decreasing.
\end{corollary}
\begin{proof}
Outside the union of false-commit events, each acceptance has $\mu_k>0$; rejection preserves the state. Apply this observation successively.
\end{proof}
This corollary guarantees neither an acceptance nor a globally optimal state. It does not compare the entire future search process with the counterfactual process following a different decision. If the score, evaluation distribution, or system state changes outside the certified transition, its premises must be reconsidered.

\subsection{An explicit anytime-valid certificate}
Let $\mathcal F_{k,n}=\mathcal G_k\vee\sigma(X_{k,1},\ldots,X_{k,n})$. Under the null, assume $\E[X_{k,i}\mid\mathcal F_{k,i-1}]\leq0$ on $H_k$. Choose fixed mixture weights $w_j\geq0$, $\sum_jw_j=1$, and fractions $0\leq\lambda_j<1$ before confirmation. Then
\[
 E_{k,n}=\sum_jw_j\prod_{i=1}^n(1+\lambda_jX_{k,i})
\]
is a nonnegative supermartingale under the conditional null, starting at one. Conditional Ville's inequality justifies acceptance when $E_{k,n}\geq1/\alpha_k$ at any inspected stopping time~\citep{howard2021time,waudby2024estimating}. A screen-selected fraction is also allowed if frozen before a valid confirmation stream. Choosing the largest wealth retrospectively without a valid mixture is not this procedure.

A fixed-sample alternative for $X\in[-1,1]$ is
$L_n=\bar X_n-\sqrt{2\log(1/\alpha_k)/n}$ at a sample size fixed before confirmation. Repeatedly checking this bound needs an additional correction or a confidence sequence. Merely increasing the sample size after an unsuccessful test does not establish validity.

\paragraph{Fixed-sample paired correctness.}
The learning experiment instead uses a one-sided exact binomial test. For $n$ prespecified IID paired correctness observations, let $W$ and $L$ count candidate-only and incumbent-only correct predictions. Conditional on $M=W+L$, the null of non-positive mean difference implies $W\sim\operatorname{Binomial}(M,q)$ with $q\leq1/2$. Thus $p=\Pr\{\operatorname{Binomial}(M,1/2)\geq W\}$ is a valid conditional $p$-value, with $p=1$ if $M=0$. Acceptance requires $p\leq\alpha_k$. This test is evaluated once at the fixed $n$; it is not an anytime-valid test. Independent uniform draws with replacement from a frozen finite pool justify this model for that pool's accuracy difference, including repeated images, but not for an external population.

\paragraph{Finite-pool reuse.}
After freezing the candidate, new independent uniform index draws from a fixed pool yield conditionally IID outcomes even when items have appeared before. Appendix~\ref{app:pool-validity} proves this for the finite-pool target; it does not justify recounting an old sample or external generalization.

\paragraph{Estimand preservation.}
Positive scaling of a raw bounded difference preserves its mean's sign. Clipping or applying a nonlinear transformation generally does not. A certificate for $\E[g(\Delta)]>0$ must be described as such; it cannot automatically support Corollary~\ref{cor:mono} for raw accuracy or return. Reporting raw paired differences alongside transformed statistics is necessary.

\section{Executable protocol and evidence costs}\paragraph{Protected verification layer.}The modifiable state comprises the candidate model, training procedure, or task code evaluated by the protocol. The evaluator, confirmation-data access, certificate implementation, and persistent budget accounting form a protected verification layer: candidate changes cannot bypass or rewrite them. The guarantees assume that this layer executes the declared rule correctly. A change to the evaluation target or verification layer requires a separate validity argument and consistent risk accounting; a certificate issued under the old rule does not authorize that change. SGM supports self-modification within this boundary, rather than unrestricted rewriting of its own verifier.
\begin{enumerate}
\item Propose and optionally screen candidate modifications using development information.
\item Select a candidate; freeze its effective configuration or checkpoint, the incumbent, the score, sampling rule, and certificate construction.
\item Open a uniquely indexed confirmation attempt and durably reserve $\alpha_k$ before revealing its observations.
\item Collect valid paired evidence and inspect it according to the specified stopping rule. Commit only on a valid crossing and matching state bindings; otherwise retain the incumbent.
\item Record the outcome and consumed allocation, including failed or interrupted attempts. Continue proposal generation from the actual resulting state.
\end{enumerate}
Multiple candidates in one outer round require separate attempts or a justified joint testing rule. A restart must not silently reset the test index. These conditions connect the statistical decision to execution; ordinary transactions and hashes can implement them. They are not separate statistical innovations.

For an unbounded sequence, $\alpha_k=\delta/[k(k+1)]$ is one valid choice. For a declared maximum $K$, $\alpha_k=\delta/(kH_K)$, $H_K=\sum_{j=1}^K1/j$, is another. The original CTHS instance $6\delta/(\pi^2k^2)$ is summable but is not a finite harmonic schedule. Indexing actual confirmation attempts is ordinary spending indexed by actual tests. Comparisons of indexing should keep weights, evidence, and screening fixed.

\paragraph{Power is a separate requirement.}
A nonzero per-attempt error rate does not imply eventual failure with probability one: positive summable levels provide an immediate counterexample. Conversely, controlling global error does not ensure affordable evidence. Under conditional IID bounded observations with $\mu_k>0$, a fixed-sample Hoeffding test accepts with probability at least $1-\beta$ whenever
\[
 n>\frac{2}{\mu_k^2}
 \left(\sqrt{\log(1/\alpha_k)}+\sqrt{\log(1/\beta)}\right)^2.
\]
Indeed, with probability at least $1-\beta$, $\bar X_n\geq\mu_k-\sqrt{2\log(1/\beta)/n}$, which makes $L_n>0$ under this condition. This is a sufficient bound for this test, not a universal sample-complexity lower bound. Verification, screening, candidate construction, and lost improvement opportunities must all enter an empirical cost assessment.

\paragraph{Design guidance.}
Match the certificate to the observation model and stopping rule. Uniform $\delta/K$ is a transparent reference for a fixed horizon; a summable schedule supports an unknown horizon. Assess power at useful effects before confirmation. Appendix~\ref{app:design-guidance} gives guidance and the negative mixture comparison; CTHS has no universal power advantage.

\section{Experiments: risk, improvement, and evidence cost}
\label{sec:empirical}
We evaluate whether a statistical acceptance rule can support state-dependent improvement, and at what evidence cost. A controlled numerical workload provides known utility differences and complete state transitions. Synthetic trajectories examine sensitivity to signal strength and path dependence. A two-attempt ImageNet-100 run tests the evidence-to-state-transition chain. Historical learning observations are retained separately where their certificates remain unresolved. Exact workload definitions, sampling rules, and random seeds are given in Appendix~\ref{app:repro}.

\subsection{Controlled numerical configuration updates}
We evaluate composite Simpson quadrature on 36 analytic functions using seven interval counts, $\{8,12,16,24,32,48,64\}$. The 252 solver outputs are computed once and cached. For each of two absolute-error tolerances, utility is the fraction of the 36 tasks solved within tolerance. Thus the full table supplies exact utility differences for diagnostic evaluation, while each decision sees sampled paired correctness differences in $\{-1,0,1\}$. Task IDs are sampled independently with replacement from this declared finite workload; this is not evaluation on unseen natural tasks.

Each policy starts at 16 intervals. A proposal moves by $\pm1$ or $\pm2$ indices from its current configuration, with boundary reflection. We run 2,000 trajectories of 40 rounds per tolerance. Policies share exogenous proposal moves and sampled task IDs; after decisions diverge, their incumbents and candidates can differ. Screening uses eight task pairs. Positive screening means open confirmation, using an equal mixture of betting fractions $\{0.1,0.25,0.5,0.75\}$ inspected at 32, 128, 512, and 1,024 paired observations. This is an instance of the certificate in Section~3.3.

We compare screening-only adoption, per-test confirmation at $\alpha=0.05$, SGM with $\alpha_k=0.05/[k(k+1)]$, its equivalent standard-spending implementation, and rejection of every proposal. SGM and standard spending agree exactly on the stored trajectories. This is an implementation consistency check, not a superiority comparison. State-dependent proposals and certificate-driven transitions connect the test decisions to the evolving configuration.

\begin{table}[tbp]
\centering\small
\caption{Controlled numerical workload: 2,000 trajectories per tolerance, 40 rounds each. FWER is the percentage of trajectories with any non-improving commitment; brackets are marginal 95\% Wilson intervals. Final utility is task success percentage. Pairs include screening and confirmation observations from the cached workload, not measured GPU time.}
\label{tab:controlled}
\begin{tabular}{@{}llrrr@{}}\toprule
Tolerance & Policy & FWER (\%) [95\% CI] & Utility (\%) & Pairs \\\midrule
$10^{-3}$ & Screen only & 76.50 [74.59, 78.31] & 75.82 & 320.0 \\
$10^{-3}$ & Per-test confirmation & 0.00 [0.00, 0.19] & 77.68 & 2634.3 \\
$10^{-3}$ & SGM / standard spending & 0.00 [0.00, 0.19] & 77.66 & 2848.0 \\
$10^{-3}$ & Reject all & 0.00 [0.00, 0.19] & 47.22 & 0.0 \\
$10^{-4}$ & Screen only & 0.00 [0.00, 0.19] & 60.96 & 320.0 \\
$10^{-4}$ & Per-test confirmation & 0.00 [0.00, 0.19] & 60.96 & 858.8 \\
$10^{-4}$ & SGM / standard spending & 0.00 [0.00, 0.19] & 60.96 & 1020.7 \\
$10^{-4}$ & Reject all & 0.00 [0.00, 0.19] & 33.33 & 0.0 \\
\bottomrule\end{tabular}
\end{table}

At tolerance $10^{-3}$, SGM has zero observed false-commit trajectories compared with 76.50\% for screening-only adoption, and final utility rises from 75.82\% to 77.66\%. Zero events in 2,000 trajectories has a Wilson upper limit of approximately 0.192\%, not zero population risk. The improvement costs approximately 8.9 times as many paired evaluations. Per-test confirmation also has zero observed errors and slightly higher final utility at lower cost. At $10^{-4}$, screening-only adoption has no observed errors and the same final utility as SGM, which still costs more. These cases demonstrate a conditional risk--power--cost trade-off, not universal efficiency or an empirical advantage of global spending over per-test confirmation.

The small, fully enumerated workload makes this a mechanism benchmark: a policy given the entire table could directly select the best configuration. Cached resampling and logical function-call counts do not represent repeated expensive solver execution or deployment of an AI Scientist.

\subsection{Sensitivity to path dependence}
We evaluate four synthetic settings combining strong-effect sizes 0.10 and 0.25 with path strengths 0 and 0.2, each with 2,000 trajectories of 40 rounds. The candidate distribution includes beneficial, zero-effect, and harmful changes; confirmation observes bounded noisy effects. The shared protocol uses four screening observations, betting fraction 0.2, and inspection times 32, 128, and 512. Utility is synthetic and dimensionless.

Across these settings SGM's observed trajectory FWER is 0.15--0.20\%, versus 3.35--3.95\% for per-test confirmation. However, SGM has lower final and cumulative utility and greater verification cost in all four comparisons. Per-test FWER does not empirically exceed 5\% in these settings; this observation does not supply it with a general trajectory guarantee. Standard spending and SGM again agree exactly. The FWER difference is driven primarily by zero-effect acceptances. SGM makes no strictly harmful commitments in these runs; per-test confirmation averages zero or 0.0005 strictly harmful commitments per trajectory, depending on the setting. Thus lower FWER here does not establish a substantial reduction in performance losses. Controlling non-improving commitments can sacrifice useful updates, and these results do not establish a universal utility benefit.

\paragraph{Error composition and missed improvements.}
A post-hoc decomposition of the stored trajectories clarifies the meaning of these risk differences (Appendix~\ref{app:composition}). Approximately 98.7--100\% of the synthetic FWER reduction from per-test confirmation to SGM comes from trajectories with zero-effect errors but no harmful commitment. Per-test confirmation has only zero or one harmful trajectory per 2,000 runs in each setting. In contrast, at numerical tolerance $10^{-3}$, every false-commit trajectory under screening-only adoption contains a strictly harmful update. Thus the meaning of a lower FWER depends on the workload and comparator. Both confirmation policies have zero observed errors in that numerical setting.

For example, at strong effect 0.25 and path strength 0, SGM accepts 5.029 beneficial updates per trajectory versus 5.358 for per-test confirmation, using 7,135 versus 5,960 evidence pairs. Unaccepted positive proposals are counted on each policy's own path, not as counterfactual terminal regret. The experiments match rounds and per-attempt caps, not realized total expenditure. Moreover, the synthetic test fixes betting fraction $\lambda=0.2$, whose expected log-wealth increment is non-positive at gains 0.10 and 0.05. Weak-effect acceptance therefore reflects this certificate's power properties as well as the risk allocation; low power cannot be attributed to the global budget alone.

\paragraph{Conditional power diagnostic.}
To interpret the reduced acceptance of weak improvements, we separately compute confirmation power conditional on the candidate effect and opened confirmation index under the synthetic IID $\{-1,+1\}$ model. At gain $g=0.10$, the original test has power 28.738\% at $\alpha=.05$, but only 4.639\% at $\alpha_{10}=.05/110$, with unchanged inspection times and a 512-sample cap. At the latter level, the randomized Neyman--Pearson power upper bound for any level-$\alpha_{10}$ test using at most 512 observations from this model is 14.575\%. Thus, both certificate efficiency and the level--sample-cap constraint limit acceptance. These post-hoc calculations are not new trajectory experiments or a causal decomposition of terminal utility differences; Appendix~\ref{app:power-diagnostic} provides the calculation and additional diagnostics.

\subsection{A certificate-driven ImageNet-100 learning loop}\label{sec:learning}
\paragraph{Protocol.}
P2 trains a ResNet-18 from random initialization for 90 epochs at learning rate 0.1. Each proposal round trains three 20-epoch branches from copies of the actual incumbent and its optimizer state. Round 1 uses learning rates $\{0.1,0.03,0.01\}$; round 2 uses $\{0.01,0.003,0.001\}$. The unchanged rate is a continuation control. Selection uses final-epoch development accuracy, requiring improvement over both parent and control. This simple proposer supplies an execution test, not a competitive training claim.

The disjoint data roles contain 49,977 training, 13,769 development, 49,968 confirmation, and 12,804 audit images after duplicate screening and quarantine. Each attempt fixes the candidate, sample size, exact paired-binomial test, and allocation before drawing 25,000 pairs independently with replacement from the common confirmation pool. Levels are $\alpha_1=.025$ and $\alpha_2=.05/6$, with trajectory budget $\delta=.05$. Sampling targets this finite pool's accuracy, not population generalization. The predeclared two-attempt cap allows no sample extension or third attempt. Appendix~\ref{app:p2} reports training details and prior negative development results.

\begin{table}[htbp]
\centering\small
\caption{Actual learning-loop decisions. Gains compare the selected checkpoint with its own parent, in percentage points. Each confirmation uses 25,000 paired draws; $W/L$ are candidate-only/parent-only correct counts.}
\label{tab:p2decisions}
\begin{tabular}{@{}lrrrrl@{}}\toprule
Attempt & Dev. gain & Conf. gain & $W/L$ & $p$ & State action \\\midrule
1 ($\alpha=.025$) & $+11.693$ & $+11.948$ & $3941/954$ & $2.00\!\times\!10^{-427}$ & Commit \\
2 ($\alpha=.00833$) & $+0.428$ & $-0.016$ & $401/405$ & $0.5699$ & Retain \\\bottomrule
\end{tabular}
\end{table}

\paragraph{Accept, commit, continue, reject.}
Attempt 1 accepts the 0.01 branch (Table~\ref{tab:p2decisions}); a separate state transition installs its full checkpoint as the incumbent. The committed weight hash matches the round-2 parent, and frozen training code and execution records document optimizer inheritance. Round 2 selects the 0.001 branch, but confirmation does not pass. The terminal incumbent remains the first committed model, with consumed allocation $1/30$. This is one accepted update followed by a rejected candidate generated from the adopted state.

\paragraph{Audit, costs, and scope.}
The initial model and all six branches are fixed before a one-time census of the 12,804 audit images. The initial and committed models score 65.36\% and 77.35\%, respectively ($+11.988$ points). The rejected candidate scores 77.39\%, only $0.039$ points above its parent. These descriptive audit outcomes do not change the incumbent or select another model. All seven outcomes and measured costs appear in Appendix~\ref{app:p2}: 210 training epochs take 4.09 hours of logged training/development time on one RTX 4090D, confirmation inference takes 31.43/31.88 seconds, and the final audit takes 115.75 seconds. These components have different timing boundaries and are not a complete research-cost total.

The run demonstrates acceptance of an improvement relative to a simple parent and refusal to commit on insufficient evidence despite a positive development gain. It does not establish that the second candidate is harmful; per-test confirmation at .05 would also reject it. One trajectory does not estimate long-horizon FWER, certify a training procedure across seeds, or show an advantage over equivalent standard spending. Earlier learning trajectories with unresolved certificates remain in Appendices~\ref{app:reported} and~\ref{app:figures}.

\section{Discussion and limitations}
SGM changes the evidence required to authorize a modification. A successful empirical demonstration must show both that the gate can accept useful updates and that its evidence model remains valid as the system evolves. The framework is useful only within that declared model; it cannot repair a misspecified objective, contaminated confirmation data, or a changed evaluator by budget accounting alone.

The formal result concerns non-improvement in the certified target. It does not bound catastrophic tail losses, guarantee lower total compute, ensure progress, or restore the classical G\"odel machine's global future-utility result. Monotonicity additionally requires a common utility and faithful state transitions. Distribution changes require a new valid sampling argument; partitioning time into epochs and assigning budgets does not itself solve distribution shift.

The controlled workload demonstrates certificate-driven configuration changes with known utility. P2 adds an actual learning-state transition and a subsequent decision, but only for one initialization, two attempts, and a restricted proposer. Its certificate concerns a fixed finite pool, and its held-out audit does not estimate population risk. Longer independently repeated trajectories would be needed to assess error calibration and retained improvements in learning systems. Engineering checks support execution conditions without establishing a separate statistical contribution.

\paragraph{A boundary of repeated fresh-evidence certification.}
Lifetime false-commit control does not itself imply eventual stagnation. A narrower limitation holds for per-attempt certification with pathwise summable levels and a uniform finite bound on the likelihood ratio of each fresh evidence transcript relative to a valid null reference. Proposition~\ref{prop:finite-acceptance} shows that the expected total number of acceptances is finite, hence infinitely many acceptances have probability zero. The argument applies to generic online testing and is included as a boundary result, not a self-modification-specific contribution. It sharpens the connection to the alpha-death discussion of \citet{feng2021approval}. Bounded sample counts alone do not imply the likelihood-ratio condition. Increasing evidence capacity or using justified shared evidence can fall outside this regime; restarting the risk budget without a separate argument does not preserve the original lifetime guarantee.

\section{Conclusion}
Statistical G\"odel Machines provide a statistical relaxation of proof-justified self-improvement. For a specified objective, conditionally valid certificates and a summable budget control the probability of any non-improving commitment across adaptive updates. Under additional fixed-objective and state-transition conditions, this implies a high-probability non-decreasing incumbent sequence. The resulting protocol makes empirical adoption executable while exposing its assumptions, missed improvements, and evidence costs. The error decomposition shows why FWER must be interpreted alongside error severity, useful acceptances, and evidence cost. The learning run establishes an executable accept--commit--continue--reject chain. Broader practical value remains to be established beyond this short finite-pool demonstration.

\bibliography{references}
\bibliographystyle{iclr2027_conference}

\clearpage
\appendix
\raggedbottom
\section{Learning-loop reproducibility and full audit}\label{app:p2}
\paragraph{Training and data.}
The ResNet-18 classifier has 100 outputs and is initialized without pretrained weights using seed 2026091207. SGD uses momentum 0.9, weight decay $10^{-4}$, batch size 128, FP32, and four data-loader workers. Training uses random resized crops and horizontal flips; evaluation resizes to 256 pixels, center-crops to 224, and uses ImageNet normalization. No warm-up or learning-rate scheduler is used in P2. The baseline endpoint is epoch 90; branch endpoints are epoch 20. Intermediate epoch-10 checkpoints are diagnostic and are not eligible for selection. Within each proposal round the three branches share the random seed (2026091301 and 2026091302 respectively), data rules, training budget, and starting model/optimizer state; only the learning rate changes.

The four roles are disjoint in content SHA-256. The frozen r2 partition removes 77 quarantined images without replacement; a perceptual-hash review additionally inspected flagged cross-role pairs. This is a documented duplicate-screening procedure, not a proof that all semantic near-duplicates have been eliminated. The confirmation role is stored in two shards of 24,984 images but is sampled as one pool of 49,968. New index draws use operating-system-backed uniform sampling with replacement. Attempts 1 and 2 contain 19,661 and 19,624 distinct images respectively; all 25,000 draws per attempt enter the fixed-sample test. The same frozen image can appear in both attempts under the declared finite-pool sampling model.

\paragraph{Adaptation and prior negative results.}
P2 is a revised development design, chosen after an earlier r2 experiment using a scheduled-learning-rate baseline and three ten-epoch candidates produced no positive development gain. That earlier trajectory stopped without opening confirmation and is retained as a negative development result. Its 120 training epochs are separate from P2's 210 epochs. P2 is not presented as an ex-ante design untouched by prior development observations. Each P2 confirmation rule and binding is frozen before its own confirmation data are revealed; the second proposal round uses the actual first committed state. The audit model list and descriptive comparisons are frozen after confirmation, before audit predictions are opened.

\begin{table}[H]
\centering\small
\caption{All seven models fixed before final audit. Dev. and audit accuracies are percentages. Round-2 branches start from the actual round-1 committed model. The audit is a census of 12,804 held-out images and is not used for model selection.}
\label{tab:p2audit}
\begin{tabular}{@{}llrrl@{}}\toprule
State & Learning rate & Dev. & Audit & Role \\\midrule
Initial & 0.1 & 64.587 & 65.362 & Initial incumbent \\
Round 1 & 0.1 & 67.064 & 68.760 & Continuation control \\
Round 1 & 0.03 & 73.520 & 74.274 & Unselected branch \\
Round 1 & 0.01 & 76.280 & 77.351 & Committed incumbent \\
Round 2 & 0.01 & 75.605 & 76.664 & Continuation control \\
Round 2 & 0.003 & 76.629 & 77.382 & Unselected branch \\
Round 2 & 0.001 & 76.709 & 77.390 & Selected, not committed \\\bottomrule
\end{tabular}
\end{table}

The predeclared audit comparisons have win/loss counts of 2,001/466 for the first candidate versus the initial model; 224/219 for the second selected candidate versus its committed parent; 1,678/578 for the first selected candidate versus its continuation control; and 368/275 for the second selected candidate versus its continuation control. The corresponding differences are 11.988, 0.039, 8.591, and 0.726 percentage points. These comparisons distinguish the effect of continued training from the candidate change without making a proposer-strength claim. No confidence interval or new acceptance test is attached to this descriptive audit.

\begin{table}[H]
\centering\small
\caption{Recorded P2 costs on one RTX 4090D. Epoch timers include training and development evaluation; confirmation timers cover paired inference only. These are distinct measured components, not total billed runtime.}
\label{tab:p2cost}
\begin{tabular}{@{}lrr@{}}\toprule
Component & Work & Seconds \\\midrule
Initial baseline & 90 epochs & 6288.64 \\
Round-1 branches including control & $3\times20$ epochs & 4199.32 \\
Round-2 branches including control & $3\times20$ epochs & 4231.15 \\
Confirmation 1 & 25,000 paired draws & 31.43 \\
Confirmation 2 & 25,000 paired draws & 31.88 \\
Final audit, all fixed models & $7\times12,804$ predictions & 115.75 \\\bottomrule
\end{tabular}
\end{table}

Epoch timers exclude checkpoint writes, preflight verification, inter-job delays, and prior development experiments. Confirmation times exclude image verification, model loading, and reservation overhead. Audit elapsed time includes image verification, loading and inference; model loading plus inference alone sums to 112.84 seconds. We do not estimate an end-to-end speedup from these non-identical timing boundaries.

\paragraph{Evidence and state binding.}
Stored records include frozen manifests and hashes, effective configurations, all epoch logs and branch checkpoints, confirmation reservations, sampled indices, raw paired outcomes, and both decisions. The first accepted checkpoint is copied into the incumbent together with optimizer state; its hash is checked before round-2 training. After the rejected second attempt, a terminal record preserves that same incumbent and records both consumed allocations. Audit predictions are stored for every model and image. Independent local checks reproduce labels, hashes, accuracies, paired counts and terminal bindings. The attempt-1 binomial tail underflows in ordinary floating-point evaluation; its nonzero value $2.004275\times10^{-427}$ is recomputed with high-precision arithmetic, without modifying the original raw result. Software is PyTorch 2.1.2+cu118, torchvision 0.16.2+cu118, and SciPy 1.13.1 for confirmation. The artifacts make this short execution trace auditable; they do not establish that a rejected candidate is harmful or that all possible future modifications satisfy the sampling assumptions.

\section{Reported learning trajectories and certificate scope}\label{app:reported}
\begin{figure}[tbp]
\centering
\includegraphics[width=\linewidth]{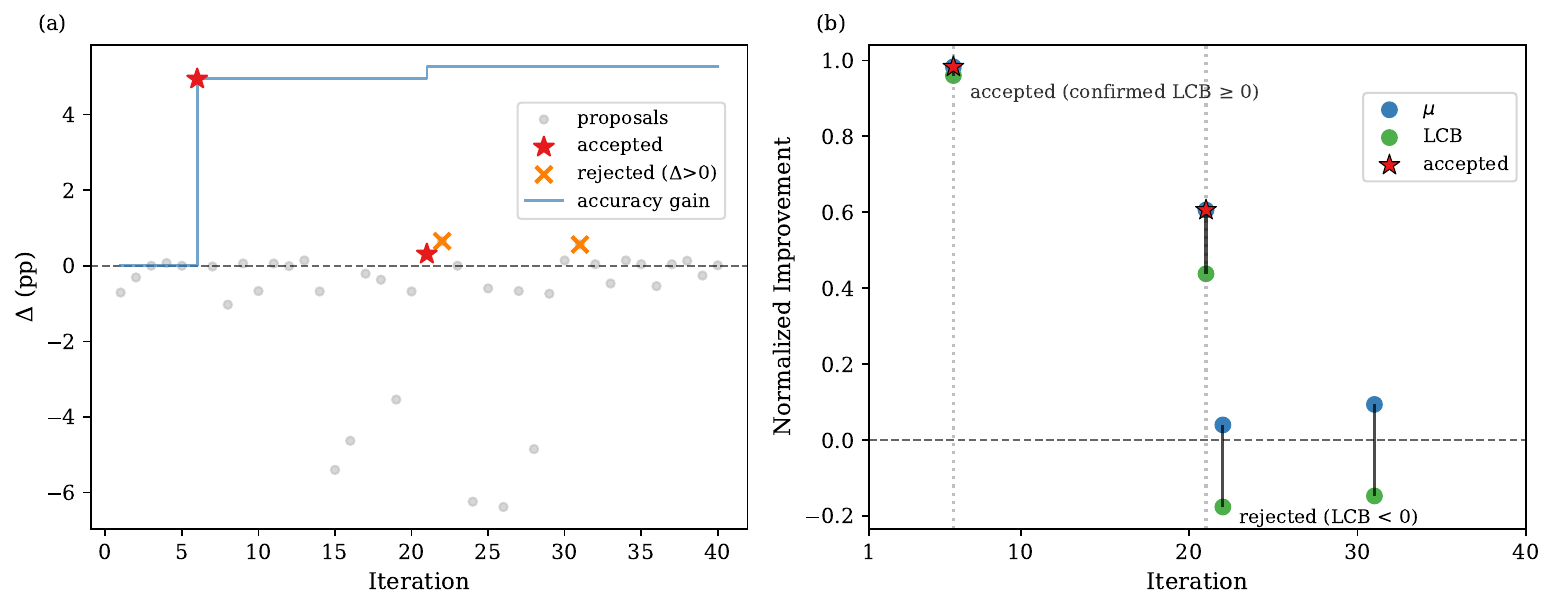}
\caption{Exploratory ImageNet-100 trajectory. (a) Reported proposal gains and incumbent gain, with adoptions at rounds 6 and 21. (b) Selected normalized estimates and historical lower-bound annotations. These annotations are not independently re-certified here; the current protocol requires a strict positive certificate for its specified estimand.}
\label{fig:long_horizon}
\end{figure}

The ImageNet-100 record reports a 40-proposal trajectory with adopted changes at rounds 6 and 21 (Figure~\ref{fig:long_horizon}). We retain this as an exploratory learning trajectory. These observations do not by themselves validate the confirmation test or the subsequent state transition.

Appendix~\ref{app:historical} gives a conditional check of the historical round-21 bound. The trajectory is not used to estimate false-commit risk or to validate the global guarantee.

For CIFAR-100, a matched 30-pair record yields a mean gain of 1.538 percentage points; the earlier 5.51-point difference uses a different baseline comparison and is not this paired effect. CIFAR-10 and LunarLander positive-effect observations also require separating reported adoption from valid certification: the supplied graphics do not support their acceptance labels with the displayed bounds. We therefore do not count these cases as additional certified improvements. Supplementary graphics and their interpretation appear in Appendix~\ref{app:figures}.

Evidence-cost comparisons use the controlled workload, where the number of observations and decision rules are specified consistently. Historical learning-cost estimates are discussed separately in Appendix~\ref{app:historical}.

\section{Supplementary figures from the reported experiments}\label{app:figures}
The following historical observations illustrate the distinction between positive estimates and statistical certification. These cases are exploratory; the controlled workload and P2 provide the certificate-driven evaluations.

\subsection{Classification and screening}
\begin{figure}[H]
\centering
\includegraphics[width=\linewidth]{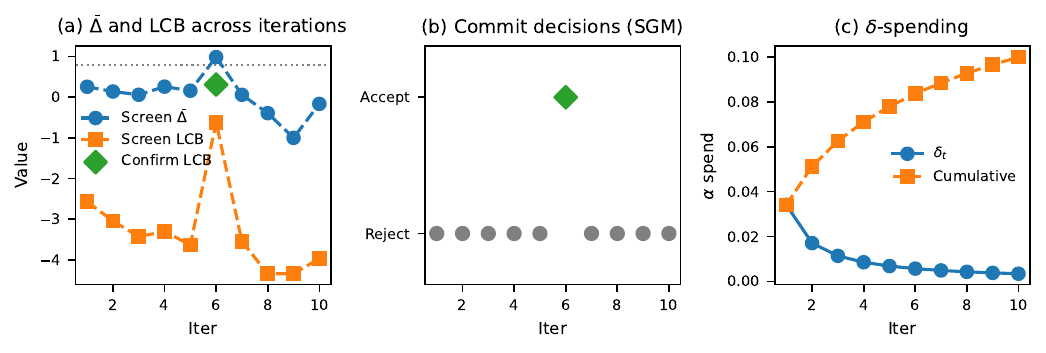}
\caption{Historical CIFAR-100 graphic. (a) Screening estimates and screening lower bounds, with a positive confirmation bound at iteration 6. (b) Reported decisions. (c) A schematic spending curve. Because the cumulative curve increases at every displayed iteration, this panel should not be read as a measured ledger of confirmation-only spending. CTHS is an optional allocation rule.}
\label{fig:cifar100}
\end{figure}

\begin{figure}[H]
\centering
\includegraphics[width=.70\linewidth]{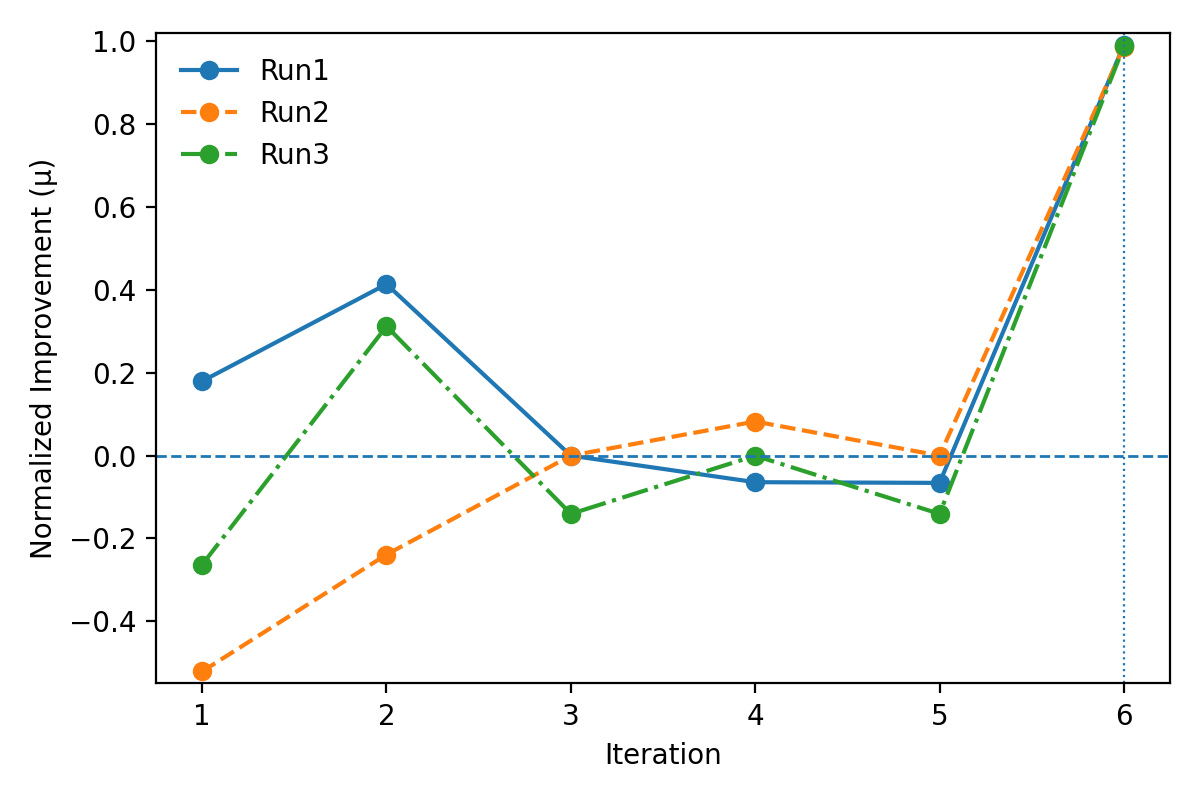}
\caption{Normalized improvement over the first six ImageNet-100 iterations in three reported runs. All three curves rise near one at iteration 6. Point estimates alone do not establish acceptance; the corresponding confirmation bounds and decision records are required.}
\label{fig:imagenet_runs}
\end{figure}

Figure~\ref{fig:cifar100} distinguishes screening from confirmation, while Figure~\ref{fig:imagenet_runs} shows the first six iterations of three ImageNet-100 runs. The latter is a plot of normalized improvement, rather than a plot of confidence bounds.

\clearpage
\subsection{Reinforcement-learning examples}
\begin{figure}[H]
\centering
\includegraphics[width=\linewidth]{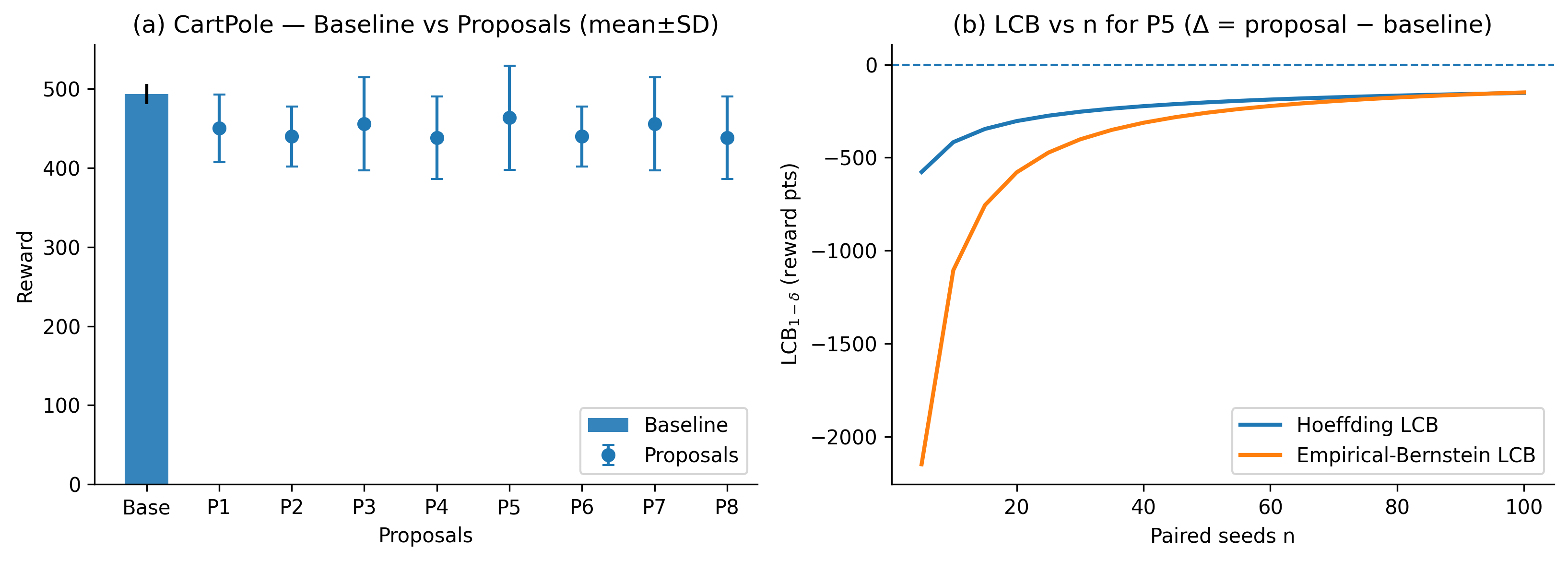}
\caption{CartPole graphic. (a) Baseline and proposal returns, with error bars labelled mean $\pm$ standard deviation in the source graphic (not 95\% confidence intervals). (b) Displayed Hoeffding and empirical-Bernstein lower bounds for P5 remain negative over the plotted sample sizes.}
\label{fig:cartpole}
\end{figure}

\begin{figure}[H]
\centering
\includegraphics[width=\linewidth]{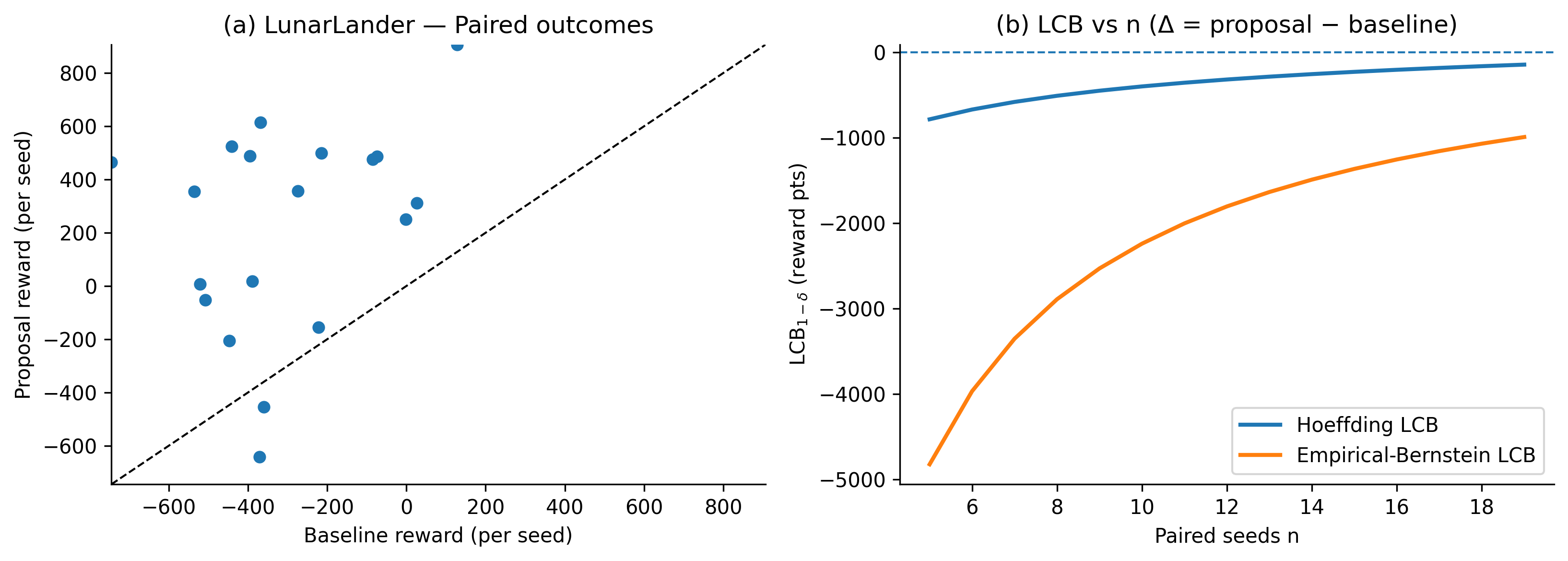}
\caption{LunarLander graphic. (a) Paired baseline and proposal returns. (b) The displayed raw-return lower bounds remain below zero up to 19 paired seeds. This graphic therefore does not demonstrate certification by these bounds; any separately reported acceptance requires its own identified estimand and test.}
\label{fig:lunarlander}
\end{figure}

Figure~\ref{fig:cartpole} illustrates proposals below a strong incumbent. Figure~\ref{fig:lunarlander} contrasts positive paired effects with conservative raw-return bounds; neither displayed bound crosses zero over the plotted sample sizes.

\clearpage
\subsection{Positive estimates without certification}
The historical CIFAR-10 paired means are positive, but the corresponding displayed bounds do not certify improvement. We report the numerical comparison below and do not count these cases as certified acceptances.

\begin{figure}[H]
\centering
\begin{tabular}{rrrr}\toprule Paired seeds & Mean gain & Hoeffding LB & EB LB \\\midrule 25 & $+2.35$ & $-46.60$ & $-79.66$ \\ 30 & $+2.22$ & $-42.47$ & $-66.12$ \\\bottomrule\end{tabular}
\caption{Historical CIFAR-10 numerical comparison, in accuracy percentage points. Values reproduce the reported means and lower-bound annotations; the empirical-Bernstein (EB) construction is not revalidated here. All displayed lower bounds are negative, so neither case is certified by these bounds. The conflicting acceptance labels in the original graphic are not used.}
\label{fig:cifar10_source}
\end{figure}

\clearpage\section{Reproducibility details}\label{app:repro}\paragraph{Numerical workload.}All integrals are over $[0,1]$. The 32 sinusoidal tasks are $f(x)=\sin(2\pi\nu x+\phi)$ for every combination of $\nu\in\{0.5,1.25,2.75,5.25,9.5,17.25,33.5,65.25\}$ and $\phi\in\{0,0.4,1.2,2.4\}$. Four further tasks use $f(x)=x^d$ for $d\in\{2,4,6,8\}$. Their exact integrals are $[\cos\phi-\cos(2\pi\nu+\phi)]/(2\pi\nu)$ and $1/(d+1)$. Composite Simpson quadrature uses the seven even interval counts in Section~5.1. Success means absolute error at most the specified tolerance. The target distribution is uniform over these 36 tasks. A proposal step is uniform over $\{-2,-1,1,2\}$; an out-of-range step is reversed. Screening and confirmation task IDs are drawn independently with replacement, separately from proposal moves. The generator is NumPy's default random generator with seed 2026091601, reset for each tolerance. Cross-policy shared draws provide paired comparisons, not additional independent trajectories.\paragraph{Synthetic workload.}Initial utility is $u=0$. Given utility $u$, the strong-gain probability is $p(u)=\operatorname{clip}(0.25+b\tanh u,0.05,0.45)$, where $b\in\{0,0.2\}$. Candidate gain is $a\in\{0.10,0.25\}$ with probability $p(u)$, $0.05$ with probability $0.20$, zero with probability $0.30$, and $-0.10$ with the remaining probability. Conditional on gain $g$, each independent screening or confirmation observation equals $+1$ with probability $(1+g)/2$ and $-1$ otherwise. Acceptance changes utility to $u+g$; rejection preserves $u$. This unbounded accumulated synthetic utility is a decision-process diagnostic, not the bounded task-success score of the numerical workload. The conditional test controls non-positive gain under the stated observation model. Each setting uses NumPy's default generator with seed 2026091301, and shares exogenous draws across policies. Global-spending policies use $\delta=0.05$ and count every opened confirmation. The simulation measures local non-improvement, not a formal comparison of entire counterfactual future utilities.\paragraph{Outcome definitions and uncertainty.}FWER counts trajectories with at least one accepted $g\leq0$; strict harm uses $g<0$. Table~\ref{tab:synthetic_detail} separates the two. Utility is averaged at the final round, cumulative utility sums the post-decision values across rounds, and observation cost includes screening and confirmation. Confidence intervals for event rates are marginal 95\% Wilson intervals across the 2,000 trajectories within each setting. These are descriptive intervals, not simultaneous coverage across all settings.\begin{table}[H]\centering\small\caption{Synthetic trajectory outcomes. FWER is in percent with marginal 95\% Wilson intervals. Harm is the mean number of strictly harmful commitments per trajectory, not a percentage.}\label{tab:synthetic_detail}\begin{tabular}{rrlrr}\toprule $a$ & $b$ & Policy & FWER (\%) [95\% CI] & Harm \\\midrule0.25 & 0 & Per-test & 3.95 [3.18,4.90] & 0.0005 \\0.25 & 0 & SGM & 0.15 [0.05,0.44] & 0 \\0.25 & 0.2 & Per-test & 3.35 [2.65,4.23] & 0 \\0.25 & 0.2 & SGM & 0.20 [0.08,0.51] & 0 \\0.10 & 0 & Per-test & 3.95 [3.18,4.90] & 0.0005 \\0.10 & 0 & SGM & 0.15 [0.05,0.44] & 0 \\0.10 & 0.2 & Per-test & 3.90 [3.14,4.84] & 0.0005 \\0.10 & 0.2 & SGM & 0.15 [0.05,0.44] & 0 \\\bottomrule\end{tabular}\end{table}\paragraph{Implementation and stored outputs.}The numerical and synthetic experiments are implemented in \texttt{run\_quadrature\_commit.py} and \texttt{run\_path\_dependency.py}, with settings in \texttt{quadrature\_commit\_r1.json} and \texttt{path\_dependency\_r1.json}. They are run as Python scripts from the project root under \texttt{iclr\_experiments/scripts/}. Stored numerical outputs include the solver table and every per-round state transition; synthetic outputs contain per-trajectory summary arrays and source hashes. Synthetic per-round histories require deterministic replay with the saved configuration and matching code. Cached evaluation counts are logical evidence costs, not measured repeated solver or GPU execution time.\clearpage
\section{Post-hoc decomposition of commitment outcomes}\label{app:composition}
This analysis uses the stored numerical and synthetic trajectories from Sections~5.1--5.2, with no new simulation or tuning. For each trajectory, let $N_+$, $N_0$, and $N_-$ count accepted positive-, zero-, and negative-effect updates. We use the disjoint identity
\[
\Pr(N_0+N_->0)=\Pr(N_->0)+\Pr(N_0>0,\,N_-=0).
\]
A trajectory can contain both zero-effect and harmful commits, so their separate marginal probabilities must not be added. The ``zero-only error'' category excludes harmful commits but may contain beneficial commits. Table~\ref{tab:error_composition} reports trajectory rates; Table~\ref{tab:retained_updates} reports mean counts and costs.

\begin{table}[htbp]
\centering\small
\caption{Synthetic error composition, in percent of 2,000 trajectories per setting. The last two columns are disjoint and sum to FWER. Existing marginal FWER intervals are in Table~\ref{tab:synthetic_detail}; zero observed harm is not zero population risk.}
\label{tab:error_composition}
\begin{tabular}{@{}rrlrrr@{}}\toprule
$a$ & $b$ & Policy & FWER & Zero-only error & Any harm \\\midrule
0.25 & 0.0 & Per-test & 3.95 & 3.90 & 0.05 \\
0.25 & 0.0 & SGM & 0.15 & 0.15 & 0.00 \\
0.25 & 0.2 & Per-test & 3.35 & 3.35 & 0.00 \\
0.25 & 0.2 & SGM & 0.20 & 0.20 & 0.00 \\
0.1 & 0.0 & Per-test & 3.95 & 3.90 & 0.05 \\
0.1 & 0.0 & SGM & 0.15 & 0.15 & 0.00 \\
0.1 & 0.2 & Per-test & 3.90 & 3.85 & 0.05 \\
0.1 & 0.2 & SGM & 0.15 & 0.15 & 0.00 \\
\bottomrule\end{tabular}
\end{table}

\begin{table}[htbp]
\centering\small
\caption{Synthetic acceptance and evidence cost, per-trajectory means. Unaccepted positive proposals include screening and confirmation failures on each policy's own path; the denominators can differ across policies. Utility is dimensionless.}
\label{tab:retained_updates}
\begin{tabular}{@{}rrlrrrr@{}}\toprule
$a$ & $b$ & Policy & Positive accepts & Unaccepted positive & Pairs & Final utility \\\midrule
0.25 & 0.0 & Per-test & 5.357 & 12.389 & 5960.4 & 1.30272 \\
0.25 & 0.0 & SGM & 5.029 & 12.717 & 7135.0 & 1.25385 \\
0.25 & 0.2 & Per-test & 7.535 & 14.621 & 5968.8 & 1.84768 \\
0.25 & 0.2 & SGM & 7.086 & 15.005 & 7689.4 & 1.76820 \\
0.1 & 0.0 & Per-test & 1.303 & 16.444 & 6524.0 & 0.12110 \\
0.1 & 0.0 & SGM & 0.321 & 17.426 & 6737.2 & 0.03107 \\
0.1 & 0.2 & Per-test & 1.353 & 16.894 & 6554.1 & 0.12638 \\
0.1 & 0.2 & SGM & 0.322 & 17.581 & 6749.5 & 0.03123 \\
\bottomrule\end{tabular}
\end{table}

At $a=0.25,b=0$, the paired SGM-minus-per-test differences are $-0.3285$ positive accepts (descriptive 95\% normal interval $[-0.3549,-0.3021]$), $-0.048875$ final utility ($[-0.053535,-0.044215]$), and $+1174.56$ evidence pairs ($[1145.10,1204.02]$). Pairing uses the original shared exogenous inputs across 2,000 trajectories. These are marginal descriptive intervals, not simultaneous claims across settings. All four settings, including weaker-effect cases, are retained.

\paragraph{Certificate-specific power limitation.}
For the synthetic $\{-1,+1\}$ observations of mean $g$, fixed fraction $\lambda=0.2$ gives expected log-wealth increment
\[
 d(g)=\frac{1+g}{2}\log(1.2)+\frac{1-g}{2}\log(0.8).
\]
Its values at $g=0.25,0.10,0.05$ are approximately $0.030272$, $-0.000138$, and $-0.010274$, respectively. Non-positive log drift does not invalidate Type-I control or rule out a finite-sample crossing, but this test does not provide eventual acceptance for every positive effect. Screening, finite sampling caps, and the chosen test therefore contribute to weak-effect non-acceptance alongside the global allocation. No betting parameter was changed in this post-hoc analysis.

\paragraph{Interpretation and reproducibility.}
The numerical workload at tolerance $10^{-3}$ has 76.5\% harmful trajectories under screening-only adoption and zero observed errors under either confirmation policy. Its contrast with the synthetic decomposition concerns different workloads and comparators, not an identified causal effect of changing error composition. Counts of unaccepted positive proposals do not quantify recoverable future utility; repeated opportunities and changed paths prevent that interpretation. Logical evidence counts are not complete measured execution costs, and equal per-attempt caps do not imply equal realized total budgets. Standard spending and SGM are exactly identical on all decomposed metrics. The analysis script \texttt{analyze\_commit\_composition.py} reads the saved arrays, checks their recorded hashes, verifies the disjoint identity and numerical event counts, and produces complete per-policy metrics and paired descriptive intervals without rerunning the simulations.

\subsection{Conditional power diagnostic}
\label{app:power-diagnostic}

\paragraph{Scope and observation model.}
This separate post-hoc calculation uses the synthetic IID model $X_i\in\{-1,+1\}$ with $\Pr_g(X_i=+1)=(1+g)/2$, so $\mathbb E_g[X_i]=g$. The null is $g\leq0$. We condition on a fixed candidate gain $g$ and an opened confirmation index $k$; no candidate generation, incumbent transition, or training is rerun. The four-observation screen requires a strictly positive sample mean and uses observations independent of confirmation. Confirmation is inspected only at $n\in\{32,128,512\}$, with a hard cap of 512 observations. We compare a per-test level $\alpha=.05$ with the summable allocation $\alpha_k=.05/[k(k+1)]$. These settings match the synthetic workload in Section~5.2.

\paragraph{Certificates and probability recursion.}
For $w$ positive observations among $n$ samples, the fixed-bet certificate is
\[
 E_n^{\mathrm{fixed}}(w)=(1+\lambda)^w(1-\lambda)^{n-w},
 \qquad \lambda=0.2.
\]
A separately specified comparison uses
\[
 E_n^{\mathrm{mix}}(w)=\frac{1}{6}\sum_{\lambda\in\Lambda}
 (1+\lambda)^w(1-\lambda)^{n-w},
 \qquad \Lambda=\{1/64,1/32,1/16,1/8,1/4,1/2\}.
\]
Each starts at one and accepts at the first inspection with $E_n\geq1/\alpha$. The mixture was fixed before this diagnostic was calculated; it was not selected after comparing diagnostic outcomes and does not replace the original fixed-bet results.

Let $q_n(w)$ denote probability mass at $(n,w)$ that has not previously crossed the threshold. Starting with $q_0(0)=1$, propagate
\[
 \widetilde q_n(w)=p q_{n-1}(w-1)+(1-p)q_{n-1}(w),
 \qquad p=(1+g)/2,
\]
with out-of-range entries zero. At inspection times, sum and remove mass for which $E_n(w)\geq1/\alpha$; at other times retain all mass. Summed first-crossing mass gives conditional confirmation power. This is a discrete probability recursion evaluated numerically, with floating-point rounding but no Monte Carlo sampling error.

\paragraph{Model-specific power upper bound.}
Under $g=0$, the full-sequence count $W_{512}$ is $\mathrm{Binomial}(512,1/2)$. For any fixed $g>0$, its likelihood ratio against $g=0$ increases with $W_{512}$. The randomized Neyman--Pearson test therefore allocates rejection probability to the largest counts, randomizing at the boundary to attain size $\alpha$. Its power upper-bounds any level-$\alpha$ rule using at most 512 observations from the same IID model: such a rule can be represented as a test on the full 512-observation sequence. The upper-tail rejection rule also has maximal null rejection probability at $g=0$ over $g\leq0$. This bound does not impose the original inspection schedule or quantify early-stopping savings; it is neither a general-task sample-complexity bound nor a claim of optimal expected evidence cost.

\begin{table}[htbp]
\centering
\small
\caption{Conditional confirmation power (\%) at $g=0.10$. All columns use the same level and 512-observation cap. The upper bound is specific to the stated IID model.}
\label{tab:conditional-power}
\begin{tabular}{lrrr}
\toprule
Level & Fixed bet & Fixed mixture & NP upper bound \\
\midrule
Per-test $\alpha=.05$ & 28.738 & 25.575 & 73.285 \\
Global $k=1$ & 23.574 & 17.533 & 61.965 \\
Global $k=2$ & 15.144 & 8.180 & 44.799 \\
Global $k=10$ & 4.639 & 1.698 & 14.575 \\
Global $k=40$ & 1.048 & 0.387 & 4.002 \\
\bottomrule
\end{tabular}
\end{table}

\paragraph{Screening and finite-budget interpretation.}
At $g=0.10$, the four-observation screen passes with probability
\[
 \Pr\{\mathrm{Binomial}(4,.55)\geq3\}
 =4(.55)^3(.45)+(.55)^4=0.39098125.
\]
Independence of screening and confirmation implies that proposal-level acceptance equals this probability times conditional confirmation power. For the fixed bet, it is 11.236\% at $\alpha=.05$ and 1.814\% at $k=10$. Thus screening, certificate efficiency, and the level--sample-cap constraint each affect acceptance in this fixed-candidate model.

The fixed bet's expected log increment is
\[
 d(g)=\tfrac12\log(1-0.2^2)
      +\tfrac g2\log(1.2/0.8),
\]
which becomes positive only above $g\approx0.1006794$. Nonpositive log drift does not preclude finite-time threshold crossing. Conversely, including weak-effect-sensitive mixture components does not ensure higher finite-budget power: at $g=0.10$ and $k=10$, the specified mixture attains 1.698\%, below the fixed bet's 4.639\%. The corresponding 14.575\% upper bound shows room for certificate improvement, but also rules out high power from changing the certificate alone under this level, sample cap, and observation model.

\paragraph{Verification and limits.}
The script \texttt{analyze\_confirmation\_power.py} computes the diagnostic and checks probability conservation, null acceptance no greater than the allocated level, positive-effect power below the Neyman--Pearson bound, agreement with a direct binomial sum for a single inspection, and monotonicity in the testing level. It holds the candidate effect and confirmation index fixed and does not identify the causal contribution of these factors to adaptive trajectory utility. It neither changes the original experimental thresholds nor constitutes new learning-trajectory evidence.

\section{Scope of historical learning evidence}\label{app:historical}For a conditional diagnostic, suppose the plotted round-21 mean 0.606 is the mean of IID observations in $[-1,1]$, the sample size is fixed at 10 or 12, this is the second confirmation attempt, and the total budget is 0.1. The fixed-sample Hoeffding lower bound is $-0.337/-0.255$ under $\alpha_2=0.1/(2H_{40})$, or $-0.309/-0.229$ under inverse-square spending. Neither supports acceptance under those assumptions. This calculation does not establish a non-positive true effect, identify the historical test, or exclude another pre-specified valid certificate. It cannot replace the original paired observations and execution record.\par The historical screening-cost calculation is $2136/12960=0.1648$. Its 12-pair assumption differs from the 10-seed trajectory description, so this ratio is an accounting illustration rather than a reconciled end-to-end cost measurement. The 40-round learning observations are retained as exploratory evidence; matching original paired observations, the declared test, and the actual next incumbent is necessary to establish a certificate-driven learning trajectory.
\clearpage
\section{Finite acceptance under bounded per-attempt evidence}
\label{app:finite-acceptance}

This section records a limitation of a class of repeated certification schemes. It does not extend the trajectory safety guarantee or provide a new statistical primitive. In particular, a decline of per-attempt power to zero alone would not establish the conclusion below: probabilities can tend to zero while remaining nonsummable.

\paragraph{Conditional experiment and null reference.}
Let $\mathcal G_k$ contain the history just before attempt $k$ as in Section~3.1. Conditional on a realized history $h$, let $Z_k$ be the new evidence transcript, including any randomization used in the decision. Write its actual conditional law as $\mathbb P_{k,h}$, and write the frozen decision as $a_k(h,Z_k)\in\{0,1\}$. The event $A_k$ is $\{a_k(h,Z_k)=1\}$. Past observations may determine the candidate, test, and allocation, but the decision must satisfy the null-reference requirement below at each such history. A null reference is a measurable probability kernel for the current conditional experiment; it need not be the conditional law of the entire adaptive process in a single alternative world.

\begin{proposition}[Finite acceptances under bounded likelihood ratios]
\label{prop:finite-acceptance}
Suppose $\alpha_k\geq0$ is $\mathcal G_k$-measurable and $\sum_{k\geq1}\alpha_k\leq\delta$ almost surely. For almost every actual pre-attempt history $h$, assume a null-reference law $\mathbb Q_{k,h}$ exists such that
\[
 \int a_k(h,z)\,\mathbb Q_{k,h}(dz)\leq\alpha_k(h).
\]
Assume also that a deterministic constant $C<\infty$, uniform over attempts and these histories, satisfies
\[
 \mathbb P_{k,h}\ll\mathbb Q_{k,h},\qquad
 \frac{d\mathbb P_{k,h}}{d\mathbb Q_{k,h}}(z)\leq C
 \quad\mathbb Q_{k,h}\text{-almost surely}.
\]
With unopened attempts assigned empty acceptance events, the total number of acceptances $N_A=\sum_{k\geq1}\mathbf1_{A_k}$ satisfies
\[
 \E[N_A]\leq C\delta,
 \qquad \Prob(N_A=\infty)=0.
\]
\end{proposition}

\begin{proof}
For almost every pre-attempt history,
\[
\begin{aligned}
 \Prob(A_k\mid\mathcal G_k=h)
 &=\int a_k(h,z)\frac{d\mathbb P_{k,h}}{d\mathbb Q_{k,h}}(z)
                 \,\mathbb Q_{k,h}(dz)\\
 &\leq C\int a_k(h,z)\,\mathbb Q_{k,h}(dz)
 \leq C\alpha_k(h).
\end{aligned}
\]
Taking expectations and using Tonelli's theorem gives
\[
 \E[N_A]=\sum_{k\geq1}\Prob(A_k)
 \leq C\sum_{k\geq1}\E[\alpha_k]
 =C\E\!\left[\sum_{k\geq1}\alpha_k\right]\leq C\delta.
\]
A nonnegative extended random variable with finite expectation is finite almost surely. No independence between attempts is needed.
\end{proof}

\paragraph{Instantiation in the synthetic observation model.}
Suppose every opened attempt uses at most a fixed $N$ fresh IID observations in $\{-1,+1\}$, with conditional mean $g_k\in[-1,1]$. The candidate and $g_k$ may depend on pre-attempt history. For a fixed history take the null reference to have mean zero, with the same auxiliary randomization as the actual test. A full length-$N$ sequence with $w$ positive observations has likelihood ratio
\[
 (1+g_k)^w(1-g_k)^{N-w}\leq(1+|g_k|)^N\leq2^N.
\]
A rule stopping earlier can be represented on this full sequence by ignoring unused observations. Thus any such rule with conditional null-reference size at most $\alpha_k$ satisfies the proposition with $C=2^N$. This statement allows shrinking positive effects and does not rely on a positive lower bound on improvement size. If each acceptance authorizes at most one committed update, only finitely many commits occur almost surely in an infinite extension of this regime.

\paragraph{What the result does not establish.}
The null-reference condition at improving histories is an additional property of the testing family, not a consequence of Equation~\ref{eq:valid} alone, whose restriction is active only at actual nulls. Uniform domination of the actual conditional law is also essential. A fixed sample cap need not bound likelihood ratios in other observation models, and bounded observations alone are insufficient. Lifetime FWER control does not require every possible procedure to use pathwise summable per-attempt levels. Consequently, the proposition does not imply that every lifetime-safe learning system must stop improving.

For $N=512$, the bound $2^{512}\delta$ is far too large to predict the number of practical accepts or a stopping round. The conclusion gives almost-sure eventual cessation of acceptances, not a known finite cutoff at which each test has exactly zero power. It does not estimate the learning run's stopping behavior: P2 has a predeclared two-attempt cap, and the synthetic experiments have 40 rounds. It also does not assert that proposals or learning computations cease. Changing the evidence-sharing mechanism, observation model, or allocation method requires checking the assumptions anew; none is automatically a valid remedy. The proof uses conditional change of measure and summability and remains unchanged for generic online testing. We therefore use it only to delimit the certification regime, in the context of prior alpha-death analyses.

\clearpage
\section{Conditional validity with a reused finite pool}
\label{app:pool-validity}

Fix the evaluation pool $D=(z_1,\ldots,z_m)$ and a deterministic scoring rule. At attempt $k$, let $\mathcal H_k=\sigma(D,\mathcal G_k)$ include the frozen candidate and incumbent as well as all earlier revealed item-level outcomes. Conditional on $\mathcal H_k$, the vector
\[
 d_{k,j}=r(c_k,z_j)-r(s_k,z_j),\qquad j=1,\ldots,m,
\]
is fixed, including entries the algorithm has not computed. Assume the new indices $I_{k,1},\ldots,I_{k,n}$ are IID uniform on $\{1,\ldots,m\}$ conditional on $\mathcal H_k$. Freeze the candidate, incumbent, level, and fixed sample size before these indices are revealed. Then $X_{k,i}=d_{k,I_{k,i}}$ are conditionally IID, with
\[
 \E[X_{k,i}\mid\mathcal H_k]
 =\mu_{k,D}:=\frac1m\sum_{j=1}^{m}d_{k,j}.
\]
For measurable sets $B_1,\ldots,B_n$, the joint law factors as
\[
 \Prob\!\left(\bigcap_{i=1}^{n}\{X_{k,i}\in B_i\}\mid\mathcal H_k\right)
 =\prod_{i=1}^{n}\frac1m\sum_{j=1}^{m}\mathbf1\{d_{k,j}\in B_i\}.
\]
Previously revealed values of $d_{k,j}$, or of earlier candidates' scores, do not change this factorization when the fresh-index assumption holds. Candidates may depend on past history, but not on the current confirmation indices or outcomes after freezing. This is inference with respect to repeated index sampling on a fixed population of items, not inference from newly sampled external examples.

\paragraph{Paired binary correctness.}
For $d_{k,j}\in\{-1,0,1\}$, let $p_+,p_-,p_0$ be the fractions of the pool having these respective outcomes. Conditional on $\mathcal H_k$, the counts $(W,L,n-W-L)$ are multinomial. Given $M=W+L>0$,
\[
 W\mid(M,\mathcal H_k)\sim\operatorname{Binomial}(M,q),
 \qquad q=\frac{p_+}{p_++p_-}.
\]
The finite-pool null $\mu_{k,D}=p_+-p_-\leq0$ implies $q\leq1/2$. The upper-tail binomial $p$-value evaluated at $1/2$ is therefore conditionally super-uniform under the null; take $p=1$ when $M=0$. This establishes the acceptance bound conditional on $\mathcal H_k$. Averaging over any additional conditioning on $D$ preserves the bound in Equation~\ref{eq:valid}, with the estimand explicitly defined as $\mu_{k,D}$. Cross-attempt independence of the selected models is unnecessary.

\paragraph{What this permits and what it does not.}
P2 uses fresh OS-backed uniform index draws with replacement and includes every draw in its fixed-sample count. An image can recur within or across attempts. Caching a deterministic prediction for a newly drawn index does not change the sampling law; simply recounting a previous sample without drawing new indices does. This argument fails if the candidate is chosen after inspecting the current indices, if sampling or item retention depends on observed performance without adjustment, if the model or score changes during confirmation, or if samples are drawn without replacement but analyzed using this IID model. Stochastic scoring requires its own execution-randomness assumptions. In particular, learning the finite pool through repeated feedback may improve that pool's score without improving external generalization. The guarantee is deliberately restricted to the former.

\section{Certificate design guidance and protocol comparisons}
\label{app:design-guidance}

\paragraph{Design workflow.}
First declare the adopted object, utility, and source of randomness. A checkpoint's item-level correctness and a training procedure's seed-level return require different observations. Next decide whether the sample size is fixed or whether success may be checked repeatedly. A valid fixed-sample paired binomial test is available for the former binary setting; an e-process or confidence sequence accommodates the latter when its conditional assumptions hold. Fixed-sample bounds must not be inspected repeatedly without adjustment.

Specify the horizon and allocation before confirmation. Uniform $\delta/K$ is a useful reference for a declared cap $K$. Decreasing summable schedules support an unbounded number of attempts but assign less evidence tolerance to late tests. Predictable allocations may use development history provided the pathwise budget constraint remains valid. Count failed opened confirmations, do not refund their allocation under ordinary spending, and do not increase a fixed-sample test's budget after observing its result. CTHS names an optional instance; it has no universal power advantage over other valid allocations.

Assess the proposed design at useful effect sizes and plausible noise levels using development information, separately from confirmation. Report conditional power and expected evidence cost, along with screening pass probability. Where a model-specific power upper bound is available, use it to distinguish test inefficiency from insufficient evidence capacity. For the IID two-point diagnostic at $g=.10$ and $k=10$, original power is 4.639\%, the specified mixture yields 1.698\%, and the 512-sample upper bound is 14.575\%. Thus the paper already includes an alternative mixture with a negative result. Neither adding small betting fractions nor changing a schedule guarantees a better utility--risk--cost trade-off. These calculations do not establish an optimal certifier for learning tasks. An inadequate prospective design may justify changing resources or the study plan before new confirmation; it does not justify relabeling previously rejected candidates as accepted.

\paragraph{Alternative error objectives.}
SGM's strict-improvement null is $H_0:\mu\leq0$. A noninferiority rule with tolerance $\varepsilon>0$ instead tests $H_0:\mu\leq-\varepsilon$ and can approve small deteriorations. Repeated approval under that target does not imply the monotonicity corollary and can accumulate losses, motivating comparisons with approved benchmarks as in \citet{feng2021approval}. Online FDR controls an expected proportion of erroneous rejections, whereas lifetime FWER controls the probability of any such error. A procedure controlling only FDR need not provide the stated lifetime guarantee. These are alternative scientific objectives, not power improvements under an unchanged objective. Likewise, a sequential A/B gate can use the same evidence tools; its stopping validity alone does not specify the target or cross-update error control of a persistent adoption protocol.

\begin{table}[htbp]
\centering\footnotesize
\caption{Protocol-level comparison of selected formulations. Scope differences are not performance rankings. Each guarantee requires its own sampling and selection assumptions.}
\label{tab:protocol-comparison}
\begin{tabularx}{\linewidth}{@{}>{\raggedright\arraybackslash}p{.15\linewidth}>{\raggedright\arraybackslash}X>{\raggedright\arraybackslash}X@{}}
\toprule
Framework & Target and evidence & Repeated decisions and relation to SGM \\
\midrule
HCPI~\citep{thomas2015high} & Policy return relative to a user-specified lower bound; off-policy trajectory evaluation under stationarity and coverage assumptions. & Batch and incremental improvement algorithms already address repeated policy updates. SGM does not supply a new off-policy estimator or validate the historical RL figures as certified improvements. \\
\addlinespace
Feng et al.~\citep{feng2021approval} & Acceptability relative to approved models, with prospective monitoring data; bad-approval counts and ratios. & Handles repeated model approval and bio-creep. Infinite-window count control bounds any bad approval for its error definition. SGM studies strict scalar improvement against the actual incumbent. \\
\addlinespace
Online FWER~\citep{tian2019online} & Potentially unbounded hypothesis streams with valid testing inputs and method-specific dependence assumptions. & Supplies established cross-test error control. SGM's matching spending implementation is an equivalence control, not an outperformed baseline. \\
\addlinespace
SEA~\citep{sengupta2026sea} & Agent architecture using anytime-valid gates; SGM-CS explicitly builds on SGM. & Broadens the agent loop; the safety of its endogenous distribution-shifting composition is left unproved. P2 does not benchmark that architecture. \\
\addlinespace
Safe Harness~\citep{cai2026safeharness} & Generation, finite-data certification, selection, adoption, and subsequent opportunities. & Analyzes feasibility and limits, with a task-level harness diagnostic. SGM's current experiments do not establish superiority over it. \\
\addlinespace
SGM (this paper) & Strict positive gain for a declared utility, candidate and incumbent frozen before valid confirmation. & Conditional lifetime false-commit bound; fixed-objective monotonicity; actual state adoption in a short finite-pool P2 run. No long-horizon learning calibration is claimed. \\
\bottomrule
\end{tabularx}
\end{table}

\paragraph{Interpreting repeated HCPI guarantees.}
HCPI includes an incremental algorithm, so the distinction cannot be single-step versus repeated improvement. The relevant questions are which policy-return threshold is certified, which trajectories and behavior policies support the estimate, and how error levels are handled across invocations. SGM's conditional composition can use a suitable valid policy certificate, but does not establish those off-policy conditions by itself. A familywise comparison must use the same error event and total level; repeated per-invocation guarantees alone should not be relabeled as a fixed lifetime guarantee.

\paragraph{Requirements for an empirical head-to-head study.}
Compare instantiated rules under a common utility, candidate source, allowed evidence access, and total cost definition. Keep the proposed change fixed for a gate-level comparison; for a closed-loop comparison allow each policy to continue from its actual adopted state and report resulting path differences. Equal per-attempt caps do not match realized total costs. Report any-error risk, strictly harmful errors, beneficial accepts, missed beneficial proposals on each policy's own path, and evidence cost. The protocol comparison above is not such an experiment, and no superiority claim is inferred from it.

\end{document}